\documentclass[conference]{IEEEtran}
\IEEEoverridecommandlockouts
\usepackage{cite}
\usepackage{amsmath,amssymb,amsfonts}
\usepackage{algorithmic}
\usepackage{graphicx}
\usepackage{textcomp}
\usepackage{xcolor}
\usepackage{array}
\usepackage{subcaption}
\usepackage{caption}
\usepackage{afterpage}
\usepackage{multirow}
\usepackage{multicol}
\usepackage{xcolor}
\usepackage{colortbl}  

\def\BibTeX{{\rm B\kern-.05em{\sc i\kern-.025em b}\kern-.08em
    T\kern-.1667em\lower.7ex\hbox{E}\kern-.125emX}}
\begin{document}

\title{A Hierarchical Compression Technique for 3D Gaussian Splatting Compression\\
}
\author{
    He Huang$^{\star}$ \qquad Wenjie Huang$^{\star}$ \qquad Qi Yang$^{\dagger}$ \qquad Yiling Xu$^{\star}$\qquad Zhu Li$^{\dagger}$\\
    \\
    $^{\star}$ Shanghai Jiao Tong University, $^{\dagger}$ University of Missouri-Kansas City \\
    \{huanghe0429, huangwenjie2023, yl.xu\}@sjtu.edu.cn, littlleempty@gmail.com, lizhu@umkc.edu
}


\maketitle

\begin{figure*}[t]
    \centering
    \includegraphics[width=\textwidth]{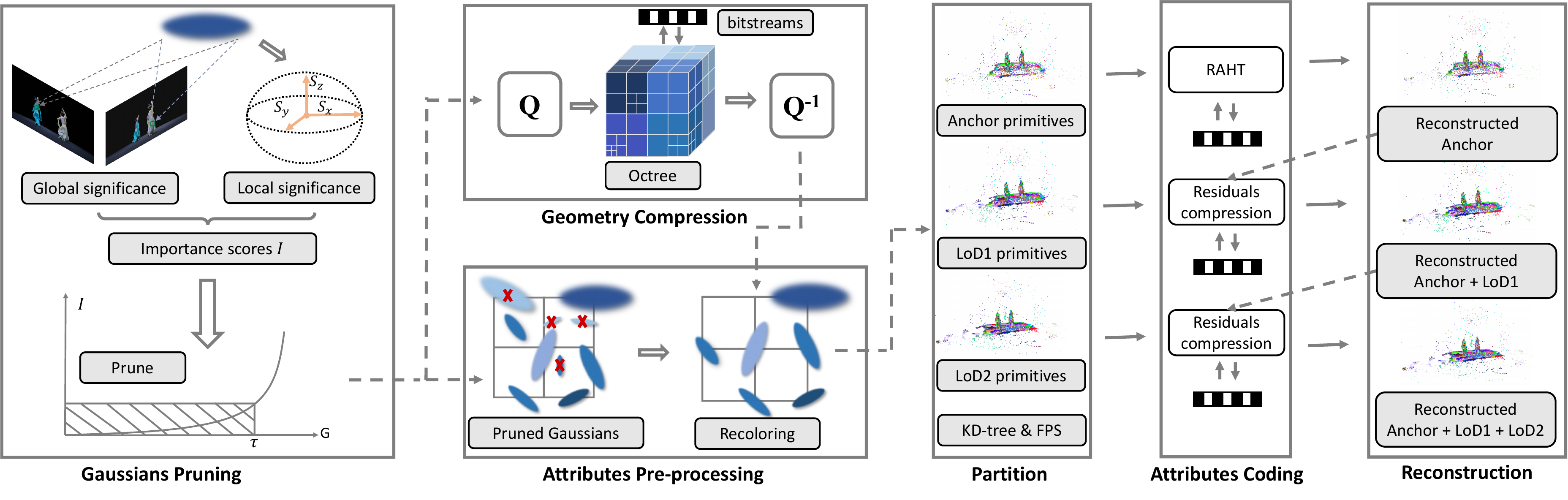}
    \vspace{-16pt}
    \caption{Framework of HGSC. \(Q\) and \(Q^{-1}\) denote the processes of quantization and dequantization, respectively.}
    \label{framework of HGSC}
    \vspace{-14pt}
\end{figure*}

\begin{abstract}
3D Gaussian Splatting (GS) demonstrates excellent rendering quality and generation speed in novel view synthesis. However, substantial data size poses challenges for storage and transmission, making 3D GS compression an essential technology. Current 3D GS compression research primarily focuses on developing more compact scene representations, such as converting explicit 3D GS data into implicit forms. In contrast, compression of the GS data itself has hardly been explored. To address this gap, we propose a Hierarchical GS Compression (HGSC) technique. Initially, we prune unimportant Gaussians based on importance scores derived from both global and local significance, effectively reducing redundancy while maintaining visual quality. An Octree structure is used to compress 3D positions. Based on the 3D GS Octree, we implement a hierarchical attribute compression strategy by employing a KD-tree to partition the 3D GS into multiple blocks. We apply farthest point sampling to select anchor primitives within each block and others as non-anchor primitives with varying Levels of Details (LoDs). Anchor primitives serve as reference points for predicting non-anchor primitives across different LoDs to reduce spatial redundancy. For anchor primitives, we use the region adaptive hierarchical transform to achieve near-lossless compression of various attributes. For non-anchor primitives, each is predicted based on the k-nearest anchor primitives. To further minimize prediction errors, the reconstructed LoD and anchor primitives are combined to form new anchor primitives to predict the next LoD. Our method notably achieves superior compression quality and a significant data size reduction of over 4.5\(\times\) compared to the state-of-the-art compression method on small scenes datasets. The code is released at https://github.com/H-Huang774/HGSC.
\end{abstract}


\section{Introduction}
3D Gaussian Splatting (GS) \cite{3DGS} has demonstrated substantial advances in the field of novel view synthesis due to its impressive visual quality with ultra fast training speed. Different from Neural Radiance Field (NeRF) \cite{nerf} with implicit representations, 3D GS uses serial explicit scattered isotropic ellipsoids to reconstruct the 3D scene. Each Gaussian consists of a 3D point center and several attributes, including scale vector, rotation quaternion, Spherical Harmonic (SH) coefficients, and opacity. Leveraging a highly optimized CUDA-based rendering implementation, 3D GS enables rapid training and rendering, making it highly suitable for practical applications. Additionally, its explicit data format is not only easy to understand and analyze, but also facilitates downstream processing (e.g., MGA \cite{streaming}), making it a promising candidate for industrial application and standardization efforts.

However, explicit point-based representations inherently result in significant storage overhead, as each point and its associated attributes must be stored independently. For example, reconstructing a large scene typically requires several million Gaussians, which can consume more than one gigabyte of memory. Therefore, compression of 3D GS becomes an essential technology to mitigate storage and transmission overhead.

Currently, research on 3D GS compression can be categorized into two distinct branches: generative compression and traditional compression \cite{GGSC}. The majority of studies focus on generative compression, employing techniques such as pruning, codebooks, and entropy constraints to produce more compact data representations during 3D GS generation. For example, LightGaussian \cite{lightgaussian} prunes insignificant Gaussians based solely on the opacity parameter, 
and HAC \cite{hac} introduces a hash-grid assisted context model to reduce spatial redundancy. However, although traditional compression has hardly been studied now, it remains an equally important area of research. GGSC \cite{GGSC} is the first work to address this gap by proposing a simple but effective graph-based compression anchor. However, the performance of GGSC is limited as it does not fully exploit the spatial redundancy of 3D GS.


In this paper, we introduce a Hierarchical GS Compression (HGSC) technique. We first prune Gaussians based on primitive importance scores assessed by both global significance and local significance: global significance is determined by each Gaussian's contribution to rendering color across different views, while local significance is associated with the volume of each Gaussian, both of which are crucial for maintaining final rendering quality. Then, an Octree \cite{octree} structure is employed to compress 3D positions. To reduce the influence of point merging within a voxel in Octree, the reconstructed points are recolored by applying the attributes of the nearest Gaussian from original 3D GS to ensure consistency.
Based on the 3D GS Octree, we implement a hierarchical compression strategy. Specifically, we use a KD-tree \cite{KD-tree} to split the 3D GS into multiple blocks and apply Farthest Point Sampling (FPS) \cite{FPS} to select anchor primitives in each block and then generate varying Levels of Details (LoDs) primitives. These anchor primitives serve as references for predicting non-anchor primitives across different LoDs to reduce spatial redundancy. For anchor primitives, we employ the region adaptive hierarchical transform (RAHT) \cite{RAHT} to achieve near-lossless compression of various attributes to enhance prediction accuracy. For non-anchor primitives, each is predicted by the k-nearest anchor primitives. Subsequently, the discrepancies between the predicted and actual attributes are quantized and subsequently encoded using the LZ77 \cite{LZ77} codec. To minimize prediction errors, the current reconstructed LoD and anchor primitives are combined to form the new anchor primitives for predicting the next LoD. Overall, our method achieves better compression efficiency and reduced processing time compared to the benchmark GGSC.

\section{Proposed Methods}

\subsection{Preliminary}
3D GS is an explicit point-based 3D scene representation through a differentiable splatting and tile-based rasterization. Each Gaussian geometry is characterized by a point center \(X\) and covariance matrix \(\Sigma\),
\setlength{\abovedisplayskip}{2pt} 
\setlength{\belowdisplayskip}{2pt} 
\begin{equation}
    G(X) = e^{-\frac{1}{2} X^{T}\Sigma^{-1}X}.
    \vspace{0pt}
\end{equation}
To maintain the positive semi-definite characteristics, \(\Sigma\) is decomposed into a scaling matrix \(S =  diag(s)\) and a rotation matrix \(R\): \(\Sigma = RSS^{T}R^{T}\).  To render an image from a random viewpoint, 3D Gaussians are first splatted to 2D planes, and the pixel value \(C \in \mathbb{R}^{3}\) is computed using \(\alpha-\)blending,
\begin{equation}
    C = \sum_{i \in N} c_i \alpha_i \prod_{j=1}^{i-1} (1 - \alpha_j),
    \label{rendering equation}
\end{equation}
where \(\alpha\) measures the opacity of each Gaussian after 2D projection, \(c \in \mathbb{R}^{3}\) is view-dependent color modeled by Spherical Harmonic (SH) coefficients, and \(N\) is the number of sorted Gaussians contributing to the rendering.

\vspace{-2pt}
\subsection{overview}
We propose an efficient hierarchical 3D GS compression technique. As depicted in Fig.~\ref{framework of HGSC}, unimportant Gaussians are initially pruned based on assessments of global and local significance. For geometry, an Octree structure is employed to compress the 3D positions. For attributes, a hierarchical compression strategy is proposed. Specifically, the anchor primitives and different LoDs are first partitioned using KD-tree and FPS techniques. Given that anchor primitives serve as references, we subsequently apply RAHT for near-lossless compression to ensure fundamental information and improve prediction accuracy. For each LoD, the attributes of each non-anchor primitives are predicted based on the k-nearest anchor primitives and lower LoDs. Subsequently, the discrepancies between the predicted and actual attributes are quantized and subsequently encoded. After completion of the compression of the current LoD, the reconstructed LoD and anchor primitives are merged to form the new anchor primitives for the subsequent LoDs, enhancing prediction accuracy. This iterative process continues until the entire 3D GS is effectively compressed.

\begin{figure}[t]
    \centering
    \begin{subfigure}[b]{0.49\linewidth}  
        \centering
        \includegraphics[width=\linewidth]{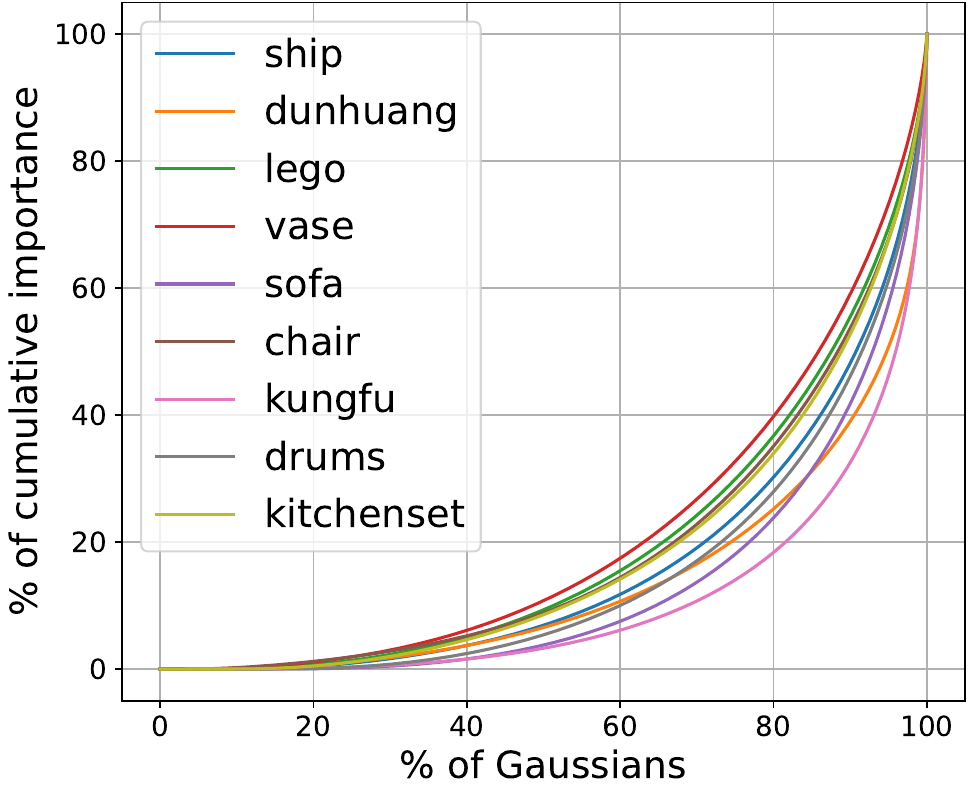}
        \caption{Small scenes}
    \end{subfigure}
    \begin{subfigure}[b]{0.49\linewidth}  
        \centering
        \includegraphics[width=\linewidth]{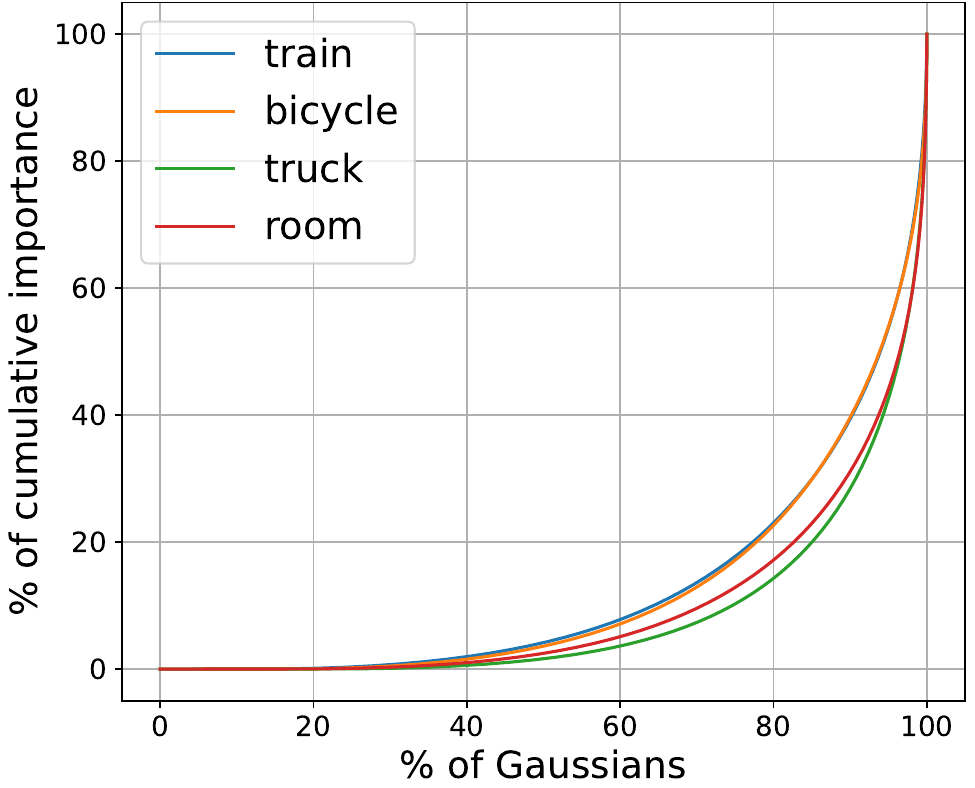}
        \caption{Big scenes}
    \end{subfigure}
    \vspace{-15pt}
    \caption{Cumulative distribution curves of importance.}
    \label{percentage curve}
    \vspace{-15pt}
\vspace{-8pt}
\end{figure}

\vspace{-8pt}
\subsection{Gaussians Pruning}
3D GS contains a large number of Gaussians and many of them are relatively insignificant for visual quality. To improve compression efficiency, we prune these less important Gaussians based on importance scores that involve two components: global significance \(I_{g}\) and local significance \(I_{l}\). The overall importance score of a Gaussian is defined as \(I = I_{g}I_{l}\), with both components being crucial for achieving high-quality final rendering. Based on \eqref{rendering equation}, we define global significance \(I_{g}\) as:
\begin{equation}
    I_{g} = \sum_{p \in \mathcal{P}} \alpha_i \prod_{j=1}^{i-1} (1 - \alpha_j) ,
\end{equation}
where \(\mathcal{P}\) represents the set of pixels that are overlapped by the projection of the Gaussian, while \(i\) denotes the set of sorted Gaussians along the ray which is determined by the \(\alpha-\)blending. The local significance \(I_{l}\) is defined as:
\begin{align}
    I_{l} = (V_{norm})^{\beta}, V_{norm} = \min \left( \max \left( \frac{V_{G}}{V_{max90}}, 0 \right), 1 \right).
\end{align}
Here, the  Gaussian volume, which is the product of the scale vector: \(V_{G} = S_{x}S_{y}S_{z}\), is firstly normalized by the 90\% largest of all sorted Gaussians. Then, it is clipped to the range between 0 and 1 to prevent excessive floating Gaussians. The \(\beta\) is introduced to provide additional flexibility.


As shown in Fig. \ref{percentage curve}, we sort the importance scores and visualize their cumulative distribution curves. Approximately 60\% of the Gaussians only account for less than 20\% of the total importance in both small and large scenes, where the importance refers to the contribution to the final rendering results. Hence, we prune the \(\tau\)\% of Gaussians with the lowest importance scores.

\subsection{Geometry Compression}
After pruning, we compress the 3D positions with Octree. The whole space is recursively divided into 8 subvoxels up to the depth of \(d\). The occupancy symbol for each voxel is composed of 8 bits (1 to 255 in decimal), where each bit indicates the occupancy status of the corresponding subvoxel. Subsequently, the mutual information among subvoxels is utilized for advanced context modeling, and each occupancy symbol is encoded using an arithmetic encoder \cite{AE}.

\subsection{Attributes Compression}

\noindent\textbf{Attributes pre-processing}. 
After geometry compression, multiple points within a voxel in Octree are merged into a single point. To maintain color consistency, the reconstructed points are recolored by applying the attributes of the nearest Gaussian from the original 3D GS. Subsequently, inspired by image compression techniques, we convert the Spherical Harmonic (SH) coefficients from the RGB to the YUV color space. This conversion is based on the principle that the Luminance Component (Y) is more critical for visual fidelity than the Chrominance components (U and V), allowing for more flexible compression parameter settings.

\noindent\textbf{Anchor primitives and different LoDs Partition}.
Based on a comprehensive visual analysis, we have identified spatial redundancy among attributes within the 3D GS. To address this, we propose sampling anchor primitives to facilitate predictions for neighboring non-anchor primitives. However, this method can be prone to significant prediction errors. Therefore, we further introduce a hierarchical compression strategy. Initially, the anchor primitives are sampled across the entire 3D GS to serve as reference points. Non-anchor primitives in higher LoDs are then predicted based on the anchor primitives and lower LoDs, which are generated using the same approach as for the anchor primitives.

Considering Octree may create numerous empty cubes due to the uneven spatial distribution of Gaussians, we employ KD-tree to divide the 3D GS into blocks, ensuring a more uniform distribution of points within each block. To ensure that the anchor primitives provide comprehensive coverage as reference points, we apply FPS within each block to select the anchor primitives. Subsequently, different LoDs are generated from the remaining non-anchor primitives by FPS across the blocks.

\begin{figure}[t]
    \centering
    \includegraphics[width=0.8\linewidth]{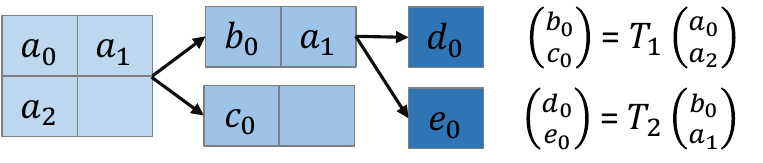}
    \caption{2D example of RAHT}
    \label{RAHT}
    \vspace{-20pt}
\end{figure}

\begin{table*}[t]
    \centering
    \caption{Quantitative results of our approach and benchmark on small and big scenes datasets. For our approach, a smaller \(\tau\) and larger quantization bit depths results in a larger size but improved fidelity, and vice versa. The best and 2nd best results are highlighted in \colorbox{red!50}{red} and \colorbox{yellow!50}{yellow} cells. E-time and D-time are short for encoding and decoding time respectively.}
    \vspace{-5pt}
    \label{scenes-comparison}
    \normalsize
    \renewcommand{\arraystretch}{1}  
    \resizebox{\linewidth}{!}{  
    \begin{tabular}{c|cccccc|cccccc}
    \hline
        \multirow{2}{*}{\textbf{Methods}} & \multicolumn{6}{c|}{\textbf{Small Scenes Datasets}} & \multicolumn{6}{c}{\textbf{Big Scenes Datasets}} \\ \cline{2-13}
        & \textbf{Size↓ (MB)} & \textbf{PSNR↑} & \textbf{SSIM↑} & \textbf{LPIPS↓} & \textbf{E-time↓ (s)} & \textbf{D-time↓ (s)} 
        & \textbf{Size↓ (MB)} & \textbf{PSNR↑} & \textbf{SSIM↑} & \textbf{LPIPS↓} & \textbf{E-time↓ (s)} & \textbf{D-time↓ (s)} \\ \hline
        3D GS \cite{3DGS} & 49.74 & - & - & - & - & - & 253.33 & - & - & - & - & - \\
        GGSC-lowrate & 16.10 & 29.20 & 0.923 & 0.072 & 119.83 & 65.14 & \cellcolor{yellow!50}42.98 & 19.06 & 0.672 & 0.357 & 381.74 & 45.35 \\
        GGSC-highrate & 20.39 & \cellcolor{yellow!50}39.95 & 0.983 & 0.024 & 383.81 & 166.08 & 78.43 & 24.24 & \cellcolor{yellow!50}0.878 & 0.167 & 1212.88 & 117.63 \\ \hline
        Ours-lowrate & \cellcolor{red!50}11.45 & 38.23 & \cellcolor{yellow!50}0.989 & \cellcolor{yellow!50}0.013 & \cellcolor{red!50}5.23 & \cellcolor{red!50}1.95 & \cellcolor{red!50}41.70 & \cellcolor{yellow!50}24.64 & 0.864 & \cellcolor{yellow!50}0.158 & \cellcolor{red!50}19.92 & \cellcolor{red!50}9.06 \\
        Ours-highrate & \cellcolor{yellow!50}15.81 & \cellcolor{red!50}41.31 & \cellcolor{red!50}0.994 & \cellcolor{red!50}0.007 & \cellcolor{yellow!50}6.37 & \cellcolor{yellow!50}2.14 & 66.85 & \cellcolor{red!50}26.00 & \cellcolor{red!50}0.897 & \cellcolor{red!50}0.118 & \cellcolor{yellow!50}49.53 & \cellcolor{yellow!50}23.97 \\ \hline
    \end{tabular}
    }
    \vspace{-15pt}
\end{table*}

\noindent\textbf{RATH for anchor primitives}.
We posit that anchor primitives are of critical importance, as they serve as reference for predicting non-anchor primitives. To preserve accuracy, RAHT is used for near-lossless compression. RAHT predicts the attributes of voxels at a higher level of the Octree by using the attributes associated with voxels from the lower level. As shown in Fig. \ref{RAHT}, RAHT merge voxels along vertical direction with transform \(T_{1}\),
\begin{equation}
    T_1 = \frac{1}{\sqrt{w_1} + \sqrt{w_2}} \begin{bmatrix} \sqrt{w_1} & \sqrt{w_2} \\ -\sqrt{w_2} & \sqrt{w_1} \end{bmatrix},
\end{equation}
where the weight coefficient \(w_i\) is the number of primitives that corresponding voxel contains. Similarly to \(T_{1}\), the coefficients \(b_0\) and \(a_1\) are merged with the transform \(T_{2}\). Ultimately, one direct current (DC) coefficient \(d_{0}\) and two alternating current (AC) coefficients \(c_{0}\) and \(e_{0}\) are acquired and encoded by arithmetic encoding \cite{AE}. The decoding process is the inverse transform. 

\noindent\textbf{Residuals compression for LoDs primitives}.
For each LoD, each primitive is predicted by the k-nearest anchor primitives. The residuals between the predicted attributes and the true attributes are quantized and then encoded by the LZ77 codec. To reduce predicted errors, the current reconstructed LoD and anchor primitives are combined as the new anchor primitives to predict the next LoD.

\afterpage{
    \begin{figure}[t]
        \centering
        \begin{subfigure}[b]{0.49\linewidth}  
            \centering
            \includegraphics[width=\linewidth]{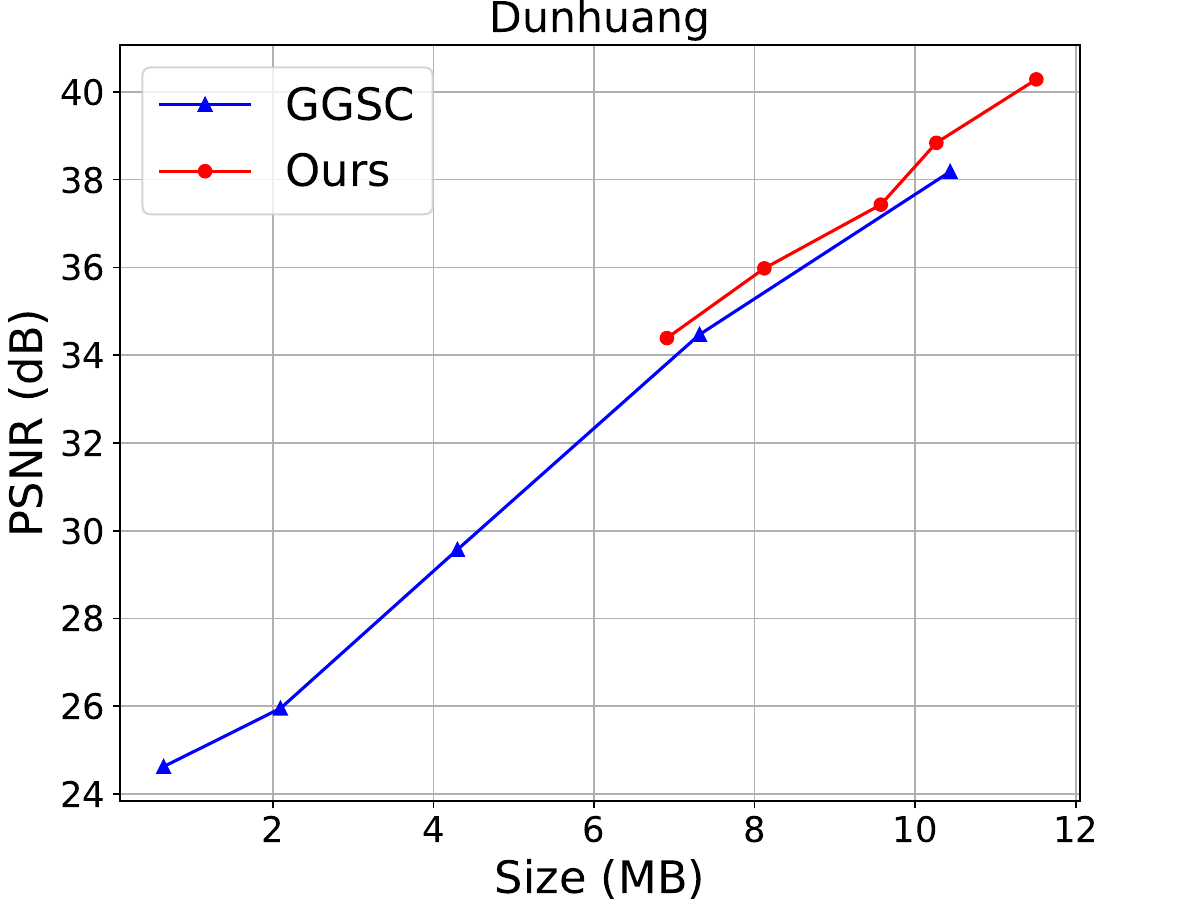}
        \end{subfigure}
        \begin{subfigure}[b]{0.49\linewidth}  
            \centering
            \includegraphics[width=\linewidth]{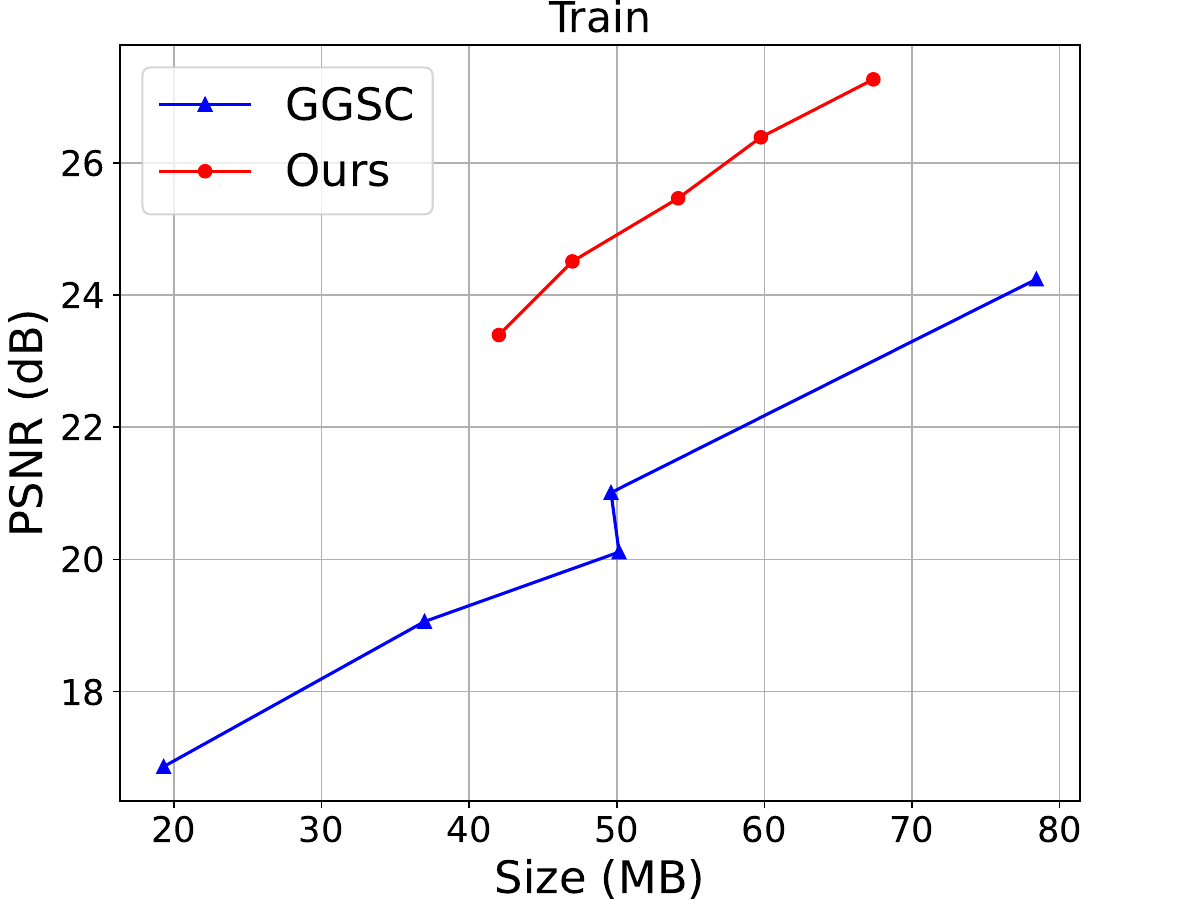}
        \end{subfigure}
        \vspace{-18pt}
        \caption{RD curves for quantitative comparisons on ``Dunhuang'' from \cite{pku} and ``Train'' from \cite{tanks}.}
        \label{RD curve}
        \vspace{-10pt}
    \end{figure}
    
    \begin{figure}[t]
        \centering
        \includegraphics[width=\linewidth]{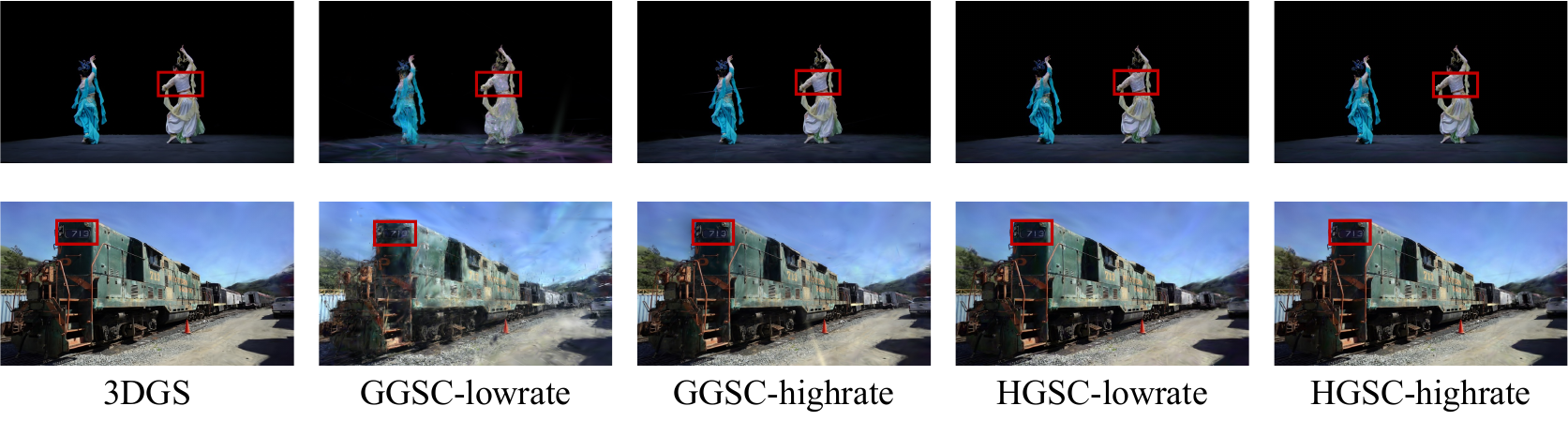}
        \vspace{-18pt}
        \captionsetup{justification=raggedright, singlelinecheck=false} 
        \caption{Qualitative comparisons of ``dunhuang'' (Top) and ``train'' (Bottom).}
        \label{Qualitative results}
        \vspace{-18pt}
    \end{figure}
}

\section{EXPERIMENT RESULTS}
\subsection{Experimental Setting}
Consistent with GGSC, we evaluate our method on two types of datasets. The small scenes often need less than one million Gaussians, including 11 different datasets from \cite{nerf, explicit_nerf_qa, pku}. Meanwhile, the big scenes require millions of Gaussians, including 4 different datasets from \cite{mip-nerf, deep-blending, tanks}. We evaluate rendering quality and compression performance using PSNR, SSIM\cite{SSIM}, LPIPS\cite{LPIPS}, size, encoding and decoding time compared to benchmark. For the hyperparameter settings, we set the depth of the Octree to 12. The pruning threshold \(\tau\)\% is set to 60\% for small scenes dataset and 66\% for big scenes dataset. Additionally, the number of LoDs is set to 2, with 10\% of the total primitives sampled as anchor primitives, 30\% for LoD1 and 60\% for LoD2.  

\subsection{Results}
The quantitative results in Tab. \ref{scenes-comparison} and the qualitative outputs in Fig. \ref{Qualitative results} clearly demonstrate the effectiveness of our proposed HGSC method. HGSC achieves a substantial size reduction of over 4.5 \(\times\) compared to the vanilla 3D GS, without any noticeable loss in rendering quality. We provide two results of different bitrate by adjusting \(\tau\) and quantization bit depths, which consistently achieve the best or second-best results across most metrics on all datasets. To further ensure a fair comparison, we present Rate-Distortion (RD) curves in Fig. \ref{RD curve}, showing that our method achieves at least a 6\% BD-rate \cite{BD-rate} improvement. This improvement can be attributed to two key factors: 1) the precise calculation of importance scores, which allows for the pruning of insignificant primitives, and 2) our hierarchical compression strategy, which effectively minimizes spatial redundancy. While GGSC manages some size reduction, it introduces substantial distortions in rendering results and leads to a considerable drop in objective quality. Moreover, our method boasts rapid decoding times, completing the process in approximately 2 seconds.

\begin{table}[t]
    \centering
    \caption{Ablation study of different stages on ``Dunhuang''. ``+'' indicates adding current model to the previous stage.}
    \vspace{-5pt}
    \normalsize
    \label{compression-techniques}
    \resizebox{\columnwidth}{!}{  
    \renewcommand{\arraystretch}{1}  
    \begin{tabular}{l|cccc}
    \hline
         \textbf{Stages} & \textbf{Size↓ (MB)} & \textbf{PSNR↑} & \textbf{SSIM↑} & \textbf{LPIPS↓} \\ \hline
          Baseline (3D GS \cite{3DGS}) & 37.75 & - & - & - \\
          + Gaussians Pruning & 14.37 & 41.20 & 0.9952 & 0.0063 \\ 
         + Geometry Compression & 13.58 & 41.08 & 0.9946 & 0.0075 \\ 
         + Attributes Compression & 8.12 & 39.78 & 0.9926 & 0.0117 \\ 
         + SH Transformation & 8.12 & 40.14 & 0.9937 & 0.0105 \\ \hline
    \end{tabular}
    \vspace{-20pt}
    }
\end{table}

\begin{table}[t]
    \centering
    \vspace{-5pt}
    \caption{Comparison of pruning strategy on ``Train''.}
    \vspace{-5pt}
    \label{compression-pruning}
    \small  
    \renewcommand{\arraystretch}{1}  
    \begin{tabular}{c|ccc}
    \hline
        \textbf{Methods} & \textbf{PSNR↑} & \textbf{SSIM↑} & \textbf{LPIPS↓}  \\ \hline
        LightGaussian \cite{lightgaussian} & 20.50 & 0.8417 & 0.1625  \\ 
        Ours ($I_{g}$) & 26.52 & 0.8976 & 0.1215  \\
        Ours ($I_{l}$) & 16.79 & 0.7452 & 0.2254  \\ 
        Ours ($I_{g}I_{l}$) & 27.26 & 0.9111 & 0.1053  \\ \hline
    \end{tabular}
    \vspace{-20pt}
\end{table}

\section{Ablation Study}
As shown in Tab. \ref{compression-techniques}, we conducted ablation studies to evaluate the impact of each component in our method. The results demonstrate that Gaussians Pruning effectively removes redundant Gaussians, retaining only those that are essential for accurate scene representation. The Geometry Compression module offers a modest reduction in size, as geometry bitstreams constitute a minor portion of the total data. Significant size reduction is achieved through Attributes Compression by employing a hierarchical compression strategy that effectively reduces spatial redundancy in the attribute data. Furthermore, the SH Transformation module improves rendering quality by transforming RGB data into the YUV domain, which is more conducive to compression. Collectively, these modules substantially enhance the overall efficiency and performance of our framework. 
In addition, as illustrated in Tab. \ref{compression-pruning}, our proposed pruning strategy outperforms LightGaussian \cite{lightgaussian}, which prunes Gaussians based solely on the opacity parameter. The results emphasize the importance of incorporating both global significance \(I_g\) and local significance \(I_l\) in the pruning process to maximize the retention of critical information. Notably, the output sizes of these strategies are consistent because they prune the same percentage of Gaussians, thereby maintaining a balanced approach to data reduction on 15 different datasets.



\section{Conclusion}

In this paper, we propose a Hierarchical 3D Gaussian Splatting Compression (HGSC) technique to enhance traditional GS compression. Initially, we prune unimportant Gaussian primitives based on both global and local significance. Then, an Octree structure is used to compress 3D positions. Following the 3D GS Octree, we implement a hierarchical compression strategy for attributes. Specifically, we use a KD-tree to split the 3DGS into multiple blocks and apply FPS to select anchor primitives in each block and non-anchor primitives within varying Levels of Details (LoDs). For anchor primitives, we employ RAHT to achieve near-lossless compression of various attributes. For non-anchor primitives, each is predicted by the k-nearest anchor. To minimize prediction errors, the current reconstructed LoD and anchor primitives are combined to form the new anchor primitives to predict the next LoD. Our method has achieved new SOTA compression performance and reconstruction rendering quality over concurrent works on 15 different datasets. Future work will focus on further reducing the time complexity.

\section*{Acknowledgment}
This paper is supported in part by National Natural Science Foundation of China (62371290, U20A20185), the Fundamental Research Funds for the Central Universities of China, and STCSM under Grant (22DZ2229005). The corresponding author is Yiling Xu(e-mail: yl.xu@sjtu.edu.cn).



\afterpage{
\bibliographystyle{IEEEtran}
\bibliography{HGSC}
}

\end{document}